\documentclass[letterpaper, 10 pt, conference]{ieeeconf}  

\IEEEoverridecommandlockouts                              

\usepackage[
    style=ieee,
    doi=false,
    isbn=false,
    url=false,
    eprint=false,
    backend=bibtex,
    natbib=true
    ]{biblatex}
\usepackage{csquotes}
\usepackage{graphicx}
\renewbibmacro*{bbx:savehash}{}

\renewcommand{\cite}{\parencite}
\usepackage[font=footnotesize]{subfig}
\usepackage{balance}
\usepackage{amsmath}
\usepackage{amsfonts}
\usepackage{balance}
\usepackage{tikz}
\usepackage{tikz-3dplot}
\usepackage{pgfplots}
\pgfplotsset{compat=1.9}
\usepgfplotslibrary{dateplot}
\usetikzlibrary{fadings}
\usepackage{pgfplotstable}
\usepackage{perl_acronyms}
\usepackage{perl_misc}
\usepackage{filecontents}
\usepackage{silence}
\usepackage{ctable}

\usepackage{marginnote}

\title{\LARGE \bf
WaterGAN: Unsupervised Generative Network to Enable Real-time Color Correction of Monocular Underwater Images
}

\author{Jie Li$^{1*}$, Katherine A. Skinner$^{2*}$, Ryan M. Eustice$^{3}$ and Matthew Johnson-Roberson$^{3}$
\thanks{$^{1}$ J. Li is with the Department of Electrical Engineering and Computer Science, University of Michigan, Ann Arbor, MI 48109 USA {\tt\small ljlijie@umich.edu}}%
\thanks{$^{2}$ K. Skinner is with the Robotics Program, University of Michigan, Ann Arbor, MI 48109 USA {\tt\small kskin@umich.edu}}
\thanks{$^{3}$ R. Eustice and M. Johnson-Roberson are with the Department of Naval Architecture and Marine Engineering, University of Michigan, Ann Arbor, MI 48109 USA {\tt\small mattjr@umich.edu}}%
\\
*These authors contributed equally to this work.
}

\begin{document}

\setlength{\parskip}{0pt plus 0.0pt minus 2.0pt}
\setlength{\floatsep}{0pt plus 1.0pt minus 0pt}
\setlength{\dblfloatsep}{0pt plus 1.0pt minus 0pt}
\setlength{\textfloatsep}{0pt plus 1.0pt minus 0pt}
\setlength{\dbltextfloatsep}{0pt plus 1.0pt minus 0pt}
\setlength{\belowdisplayskip}{2pt} \setlength{\belowdisplayshortskip}{4pt}
\setlength{\abovedisplayskip}{0pt} \setlength{\abovedisplayshortskip}{0pt}

\maketitle
\thispagestyle{empty}
\pagestyle{empty}

\begin{abstract}

This paper reports on WaterGAN, a generative adversarial network (GAN) for generating realistic underwater images from in-air image and depth pairings in an unsupervised pipeline used for color correction of monocular underwater images. Cameras onboard autonomous and remotely operated vehicles can capture high resolution images to map the seafloor; however, underwater image formation is subject to the complex process of light propagation through the water column. The raw images retrieved are characteristically different than images taken in air due to effects such as absorption and scattering, which cause attenuation of light at different rates for different wavelengths. While this physical process is well described theoretically, the model depends on many parameters intrinsic to the water column as well as the structure of the scene. These factors make recovery of these parameters difficult without simplifying assumptions or field calibration; hence, restoration of underwater images is a non-trivial problem. Deep learning has demonstrated great success in modeling complex nonlinear systems but requires a large amount of training data, which is difficult to compile in deep sea environments. Using WaterGAN, we generate a large training dataset of corresponding depth, in-air color images, and realistic underwater images. This data serves as input to a two-stage network for color correction of monocular underwater images. Our proposed pipeline is validated with testing on real data collected from both a pure water test tank and from underwater surveys collected in the field. Source code, sample datasets, and pretrained models are made publicly available.

\end{abstract}

\begin{keywords}
Underwater vision, monocular vision, generative adversarial network, image restoration
\end{keywords}

\section{Introduction}
\begin{figure}[ht!]
    \centering
    \includegraphics[width=0.95\columnwidth]{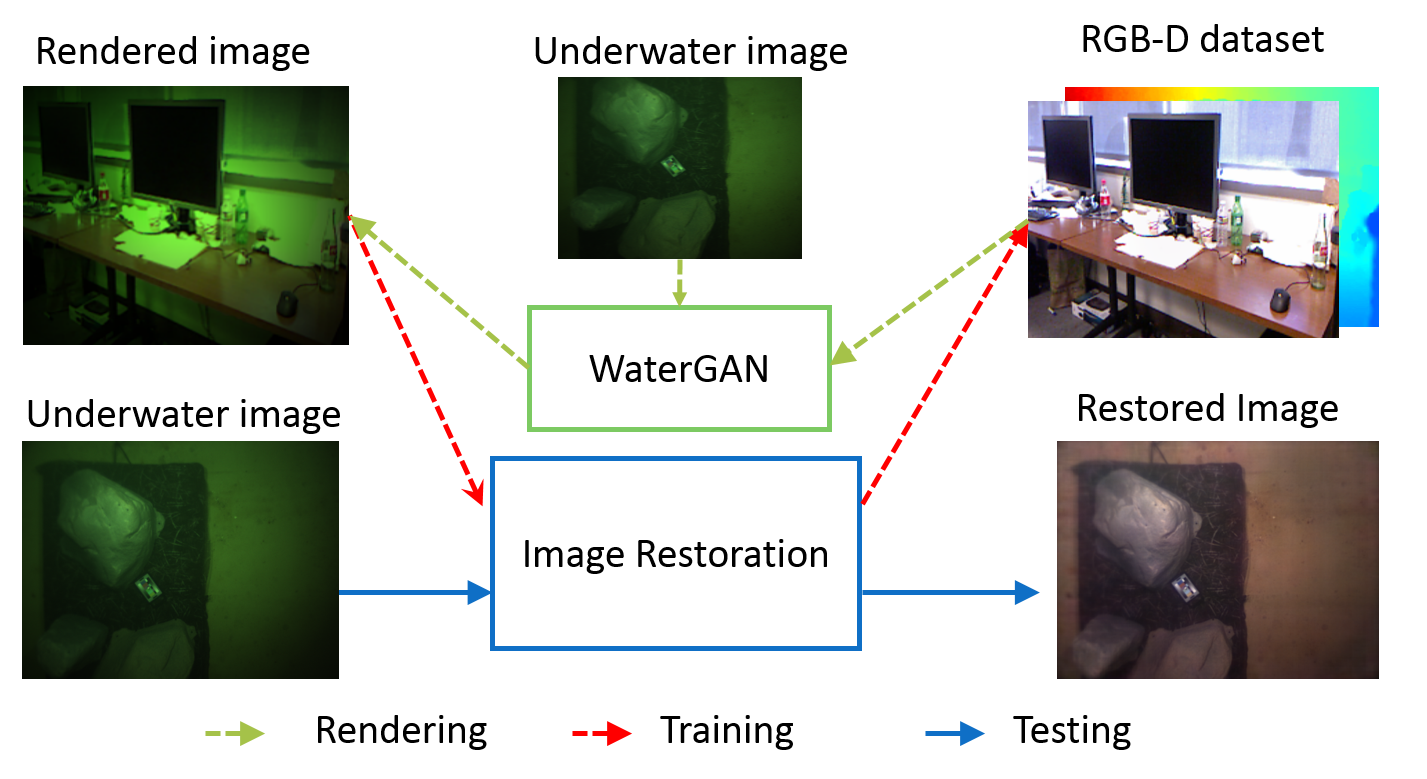}
    \caption{Flowchart displaying both the WaterGAN and color correction networks. WaterGAN takes input in-air RGB-D and a sample set of underwater images and outputs synthetic underwater images aligned with the in-air RGB-D. The color correction network uses this aligned data for training. For testing, a real monocular underwater image is input and a corrected image and relative depth map are output.}
    \label{fig:overview}
\end{figure} 

Many fields rely on underwater robotic platforms equipped with imaging sensors to provide high resolution views of the seafloor. For instance, marine archaeologists use photomosaic maps to study submerged objects and cities~\cite{Johnson-Roberson:2016aa}, and marine scientists use surveys of coral reef systems to track bleaching events over time~\cite{Bryson:2013kx}. While recent decades have seen great advancements in vision capabilities of underwater platforms, the subsea environment presents unique challenges to perception that are not present on land. Range-dependent lighting effects such as attenuation cause exponential decay of light between the imaged scene and the camera. This attenuation acts at different rates across wavelengths and is strongest for the red channel. As a result, raw underwater images appear relatively blue or green compared to the true color of the scene as it would be imaged in air. Simultaneously, light is added back to the sensor through scattering effects, causing a haze effect across the scene that reduces the effective resolution. In recent decades, stereo cameras have been at the forefront in solving these challenges. With calibrated stereo pairs, high resolution images can be aligned with depth information to compute large-scale photomosaic maps or metrically accurate 3D reconstructions~\cite{Johnson-Roberson:2013zv}. However, degradation of images due to range-dependent underwater lighting effects still hinders these approaches, and restoration of underwater images involves reversing effects of a complex physical process with prior knowledge of water column characteristics for a specific survey site.

Alternatively, neural networks can achieve end-to-end modeling of complex nonlinear systems. Yet deep learning has not become as commonplace subsea as it has for terrestrial applications. One challenge is that many deep learning structures require large amounts of training data, typically paired with labels or corresponding ground truth sensor measurements. Gathering large sets of underwater data with depth information is challenging in deep sea environments; obtaining ground truth of the true color of a natural subsea scene is also an open problem.

Rather than gathering training data, we propose a novel approach, WaterGAN, a generative adversarial network (GAN)~\cite{GAN} that uses real unlabeled underwater images to learn a realistic representation of water column properties of a particular survey site. WaterGAN takes in-air images and depth maps as input and generates corresponding synthetic underwater images as output. This dataset with corresponding depth data, in-air color, and synthetic underwater color can then supplant the need for real ground truth depth and color in the training of a color correction network. We propose a color correction network that takes as input raw unlabeled underwater images and outputs restored images that appear as if they were taken in air. 

This paper is organized as follows:~\secref{s:background} presents relevant prior work;~\secref{s:technical} gives a detailed description of our technical approach;~\secref{s:exp} presents our experimental setup to validate our proposed approach;~\secref{s:res} provides results and a discussion of these results; lastly,~\secref{s:con} concludes the paper.

\addtolength{\voffset}{0.5cm}
\section{Background}\label{s:background}

Prior work on compensating for effects of underwater image formation has focused on explicitly modeling this physical process to restore underwater images to their true color. Jordt et al. used a modified Jaffe-McGlamery model with parameters obtained through prior experiments~\cite{jordtthesis}~\cite{Jaffe:1990aa}. However, attenuation parameters vary for each survey site depending on water composition and quality. 
Bryson et al. used an optimization approach to estimate water column and lighting parameters of an underwater survey to restore the true color of underwater scenes~\cite{Bryson:2015aa}. However, this method requires detailed knowledge of vehicle configuration and the camera pose relative to the scene. In this paper, we propose to learn to model these effects using a deep learning framework without explicitly encoding vehicle configuration parameters. 

Approaches that make use of the gray world assumption~\cite{Johnson-Roberson:2016aa} or histogram equalization are common preprocessing steps for underwater images and may result in improved image quality and appearance. However, as such methods have no knowledge of range-dependent effects, resulting images of the same object viewed from different viewpoints may appear with different colors. 
Work has been done to enforce the consistency of restored images across a scene~\cite{Bryson:2012fk}, but these methods require dense depth maps. In prior work, Skinner et al. worked to relax this requirement using an underwater bundle adjustment formulation to estimate the parameters of a fixed attenuation model and the 3D structure simultaneously~\cite{Skinner:2016ab}, but such approaches require a fixed image formation model and handle unmodeled effects poorly. Our proposed approach can perform restoration with individual  monocular images as input, and learns the relative structure of the scene as it corrects for the effects of range-dependent attenuation.

Several methods have addressed range-dependent image dehazing by estimating depth through developed or statistical priors on attenuation effects~\cite{ncarlevaris-2010a,Drews_2013_ICCV_Workshops,DrewsNBC16}. More recent work has focused on leveraging the success of deep learning techniques to estimate parameters of the complex physical model. Shin et al.~\cite{yshin-2016-oceans} developed a deep learning pipeline that achieves state-of-the-art performance in underwater image dehazing using simulated data with a regression network structure to estimate parameters for a fixed restoration model. 
Our method incorporates real field data in a generative network to learn a realistic representation of environmental conditions for raw underwater images of a specific survey site.

We structure our training data generator, WaterGAN, as a generative adversarial network (GAN). GANs have shown success in generating realistic images in an unsupervised pipeline that only relies on an unlabeled set of images of a desired representation~\cite{GAN}. A standard GAN generator receives a noise vector as input and generates a synthetic image from this noise through a series of convolutional and deconvolutional layers~\cite{DCGAN}. Recent work has shown improved results by providing an input image to the generator network, rather than just a noise vector. Shrivastava et al. provided a simulated image as input to their network, SimGAN, and then used a refiner network to generate a more realistic image from this simulated input~\cite{SimGAN}. 
To extend this idea to the domain of underwater image restoration, we also incorporate easy-to-gather in-air RGB-D data into the generator network since underwater image formation is range-dependent. 
Sixt et al. proposed a related approach in RenderGAN, a framework for generating training data for the task of tag recognition in cluttered images~\cite{RenderGAN}. RenderGAN uses an augmented generator structure with augment functions modeling known characteristics of their desired images, including blur and lighting effects. RenderGAN focuses on a finite set of tags and classification as opposed to a generalizable transmission function and image-to-image mapping. 


\section{Technical Approach}\label{s:technical}
This paper presents a two-part technical approach to produce a pipeline for image restoration of monocular underwater images. Figure~\ref{fig:overview} shows an overview of our full pipeline. WaterGAN is the first component of this pipeline, taking as input in-air RGB-D images and a sample set of underwater images to train a generative network adversarially. This training procedure uses unlabeled raw underwater images of a specific survey site, assuming that water column effects are mostly uniform within a local area. 
This process produces rendered underwater images from in-air RGB-D images that conform to the characteristics of the real underwater data at that site. These synthetic underwater images can then be used to train the second component of our system, a novel color correction network that can compensate for water column effects in a specific location in real-time. 

\subsection{Generating Realistic Underwater Images}\label{subsec:gan}

\begin{figure*}[h]
    \centering
    \includegraphics[width=1.95\columnwidth]{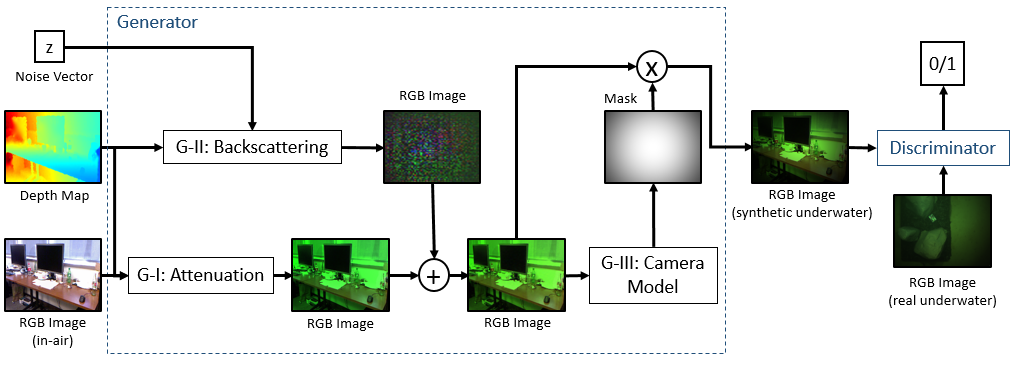}
    \caption{WaterGAN: The GAN for generating realistic underwater images with similar image formation properties to those of unlabeled underwater data taken in the field.}
    \label{fig:G}
\end{figure*}

We structure WaterGAN as a generative adversarial network, which has two networks training simultaneously: a generator, $G$, and a discriminator, $D$ (Fig.~\ref{fig:G}). In a standard GAN~\cite{GAN}~\cite{DCGAN} the generator input is a noise vector $z$, which is projected, reshaped, and propagated through a series of convolution and deconvolution layers. The output is a synthetic image, $G(z)$. The discriminator receives as input the synthetic images and a separate dataset of real images, $x$, and classifies each sample as real (1) or synthetic (0). The goal of the generator is to output synthetic images that the discriminator classifies as real. Thus in optimizing $G$, we seek to maximize

\begin{align}
log(D(G(z)).
\end{align}

\noindent The goal of the discriminator is to achieve high accuracy in classification, minimizing the above function, and maximizing $D(x)$ for a total value function of

\begin{align}
log(D(x)) + log(1 - D(G(z))).
\end{align}





The generator of WaterGAN features three main stages, each modeled after a component of underwater image formation: attenuation (G-I), backscattering (G-II), and the camera model (G-III). The purpose of this structure is to ensure that generated images align with the RGB-D input, such that each stage does not alter the underlying structure of the scene itself, only its relative color and intensity. Additionally, our formulation ensures that the network is using depth information in a realistic manner. This is necessary as the discriminator does not have direct knowledge of the depth of the scene. The remainder of this section describes each stage in detail.

\noindent \underline{G-I: Attenuation}


The first stage of the generator, G-I, accounts for range-dependent attenuation of light. The attenuation model is a simplified formulation of the Jaffe-McGlamery model~\cite{Jaffe:1990aa}~\cite{McGlamery:1975aa},

\begin{align}
G_{1} = I_{air}e^{-\eta(\lambda)r_{c}},
\label{eq:att-ref}
\end{align}

\noindent where $I_{air}$ is the input in-air image, or the initial irradiance before propagation through the water column, $r_{c}$ is the range from the camera to the scene, and $\eta$ is the wavelength-dependent attenuation coefficient estimated by the network. We discretize the wavelength, $\lambda$, into three color channels. $G_{1}$ is the final output of G-I, the final irradiance subject to attenuation in the water column. Note that the attenuation coefficient is dependent on water composition and quality, and varies across survey sites. To ensure that this stage only attenuates light, as opposed to adding light, and that the coefficient stays within physical bounds, we constrain $\eta$ to be greater than $0$. All input depth maps and images have dimensions of $48\times 64$ for training model parameters. This training resolution is sufficient for the size of our parameter space and preserves the aspect ratio of the full-size images. Note that we can still achieve full resolution output for final data generation, as explained below. 
Depth maps for in-air training data are normalized to the maximum underwater survey altitude expected. Given the limitation of optical sensors underwater, it is reasonable to assume that this value is available.

\noindent \underline{G-II: Scattering}

As a photon of light travels through the water column, it is also subjected to scattering back towards the image sensor. This creates a characteristic haze effect in underwater images and is modeled by 

\begin{align}
B = \beta(\lambda)(1-e^{-\eta(\lambda)r_{c}}),
\label{eq:scattering-ref}
\end{align}

\noindent where $\beta$ is a scalar parameter dependent on wavelength. Stage G-II accounts for scattering through a shallow convolutional network. To capture range-dependency, we input the $48\times 64$ depth map and a 100-length noise vector. The noise vector is projected, reshaped, and concatenated to the depth map as a single channel $48\times 64$ mask. To capture wavelength-dependent effects, we copy this input for three independent convolution layers with kernel size $5\times 5$. This output is batch normalized and put through a final leaky rectified linear unit (LReLU) with a leak rate of 0.2. Each of the three outputs of the distinct convolution layers are concatenated together to create a $48\times 64\times 3$ dimension mask. Since backscattering adds light back to the image, and to ensure that the underlying structure of the imaged scene is not distorted from the RGB-D input, we add this mask, $M_{2}$, to the output of G-I:

\begin{align}
G_{2} = G_{1} + M_{2}.
\label{eq:scatt2-ref}
\end{align}

\noindent \underline{G-III: Camera Model}

Lastly we account for the camera model. First we model vignetting, which produces a shading pattern around the borders of an image due to effects from the lens. We adopt the vignetting model from~\cite{vignette},

\begin{align}
V = 1 + a r^2 + b r^4 + c r^6,
\label{eq:vig-ref}
\end{align}

\noindent where $r$ is the normalized radius per pixel from the center of the image, such that $r = 0$ in the center of the image and $r = 1$ at the boundaries. The constants $a$, $b$, and $c$ are model parameters estimated by the network. The output mask has dimensions of the input images, and $G_{2}$ is multiplied by $M_3 = \frac{1}{V}$ to produce a vignetted image $G_{3}$,

\begin{align}
G_{3} = M_{3} G_{2}.
\label{eq:vig2-ref}
\end{align}

\noindent As described in~\cite{vignette}, we constrain this model by

\begin{align}
(c \geq 0) \land (4b^2 - 12ac < 0).
\label{eq:vig2-ref}
\end{align}

Finally we assume a linear sensor response function, which has a single scaling parameter $k$~\cite{Bryson:2015aa}, with the final output given by

\begin{align}
G_{out} = k G_{3}.
\label{eq:vig2-ref}
\end{align}

\noindent \underline{Discriminator}

For the discriminator of WaterGAN, we adopt the convolutional network structure used in~\cite{DCGAN}. 
The discriminator takes an input image $48\times 64\times 3$, real or synthetic. This image is propagated through four convolutional layers with kernel size $5\times 5$ with the image dimension downsampled by a factor of two, and the channel dimension doubled. Each convolutional layer is followed by LReLUs with a leak rate of 0.2. The final layer is a sigmoid function and the discriminator returns a classification label of (0) for synthetic or (1) for a real image.



\noindent \underline{Generating Image Samples}

After training is complete, we use the learned model to generate image samples. For image generation, we input in-air RGB-D data at a resolution of $480\times 640$ and output synthetic underwater images at the same resolution. To maintain resolution and preserve the aspect ratio, the vignetting mask and scattering image are upsampled using bicubic interpolation before applying them to the image. The attenuation model is not specific to the resolution.

\subsection{Underwater Image Restoration Network\label{subsec:model-learning}}
\begin{figure*}
    \centering
    \includegraphics[width=1.85\columnwidth]{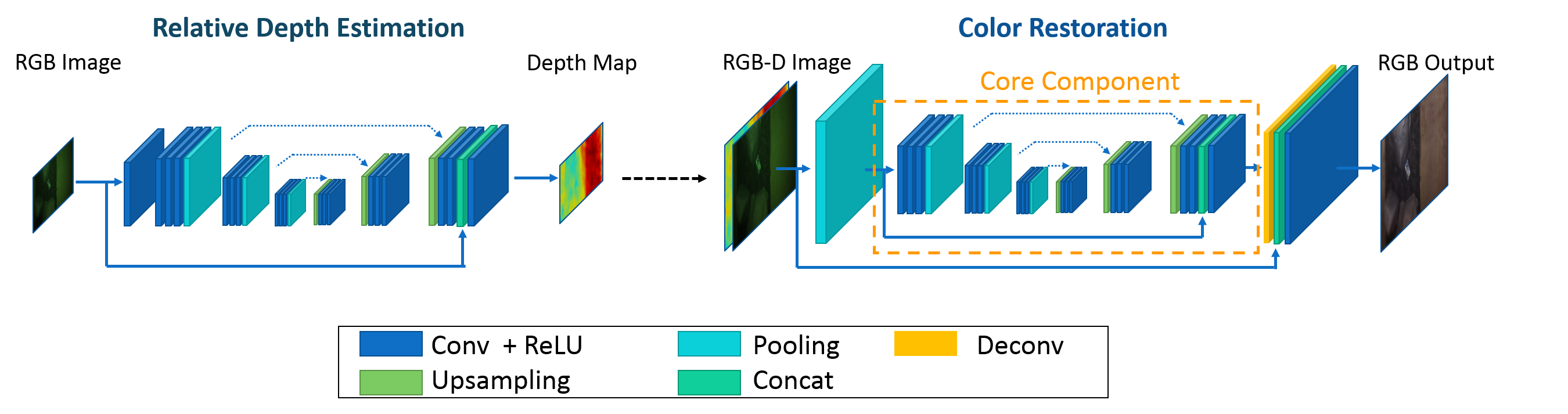}
    \caption{Network architecture for color estimation. The first stage of the network takes a synthetic (training) or real (testing) underwater image and learns a relative depth map. The image and depth map are then used as input for the second stage to output a restored color image as it would appear in air.}
    \label{fig:color-module}
\end{figure*}
To achieve real-time monocular image color restoration, we propose a two-stage algorithm using two fully convolutional networks that train on the in-air RGB-D data and corresponding rendered underwater images generated by WaterGAN. The architecture of the model is depicted in \figref{fig:color-module}. A depth estimation network first reconstructs a coarse relative depth map from the downsampled synthetic underwater image. Then a color restoration network conducts restoration from the input of both the underwater image and its estimated relative depth map. 

We propose the basic architecture of both network modules based on a state-of-the-art fully convolutional encoder-decoder architecture for pixel-wise dense learning, SegNet~\cite{segnet}. A new type of non-parametric upsampling layer is proposed in SegNet that directly uses the index information from corresponding max-pooling layers in the encoder. The resulting encoder-decoder network structure has been shown to be more efficient in terms of training time and memory compared to benchmark architectures that achieve similar performance. 
SegNet was designed for scene segmentation, so preserving high frequency information of the input image is not a required property. In our application of image restoration, however, it is important to preserve the texture level information for the output so that the corrected image can still be processed or utilized in other applications such as 3D reconstruction or object detection.
Inspired by recent work on image restoration and denoising using neural networks~\cite{mao2016image}\cite{jain2007supervised}, we incorporate skipping layers on the basic encoder-decoder structure to compensate for the loss in high frequency components through the network. The skipping layers are able to increase the convergence speed in network training and to improve the fine scale quality of the restored image, as shown in \figref{fig:detail}. More discussion will be given in \secref{s:res}.

As shown in \figref{fig:color-module}, in the depth estimation network, the encoder consists of 10 convolution layers and three levels of downsampling. The decoder is symmetric to the encoder, using non-parametric upsampling layers. Before the final convolution layer, we concatenate the input layer with the feature layers to provide high resolution information to the last convolution layer. The network takes a downsampled underwater image of $56\times56\times3$ as input and outputs a relative depth map of $56\times56\times1$. This map is then upsampled to $480\times480$ and serves as part of the input to the second stage for color correction.

The color correction network module is similar to the depth estimation network. It takes an input RGB-D image at the resolution of $480\times480$, padded to $512\times 512$ to avoid edge effects.  Although the network module is a fully convolutional network and changing the input resolution does not affect the model size itself, increasing input resolution demands larger computational memory to process the intermediate forward and backward propagation between layers. A resolution of $256\times256$ would reach the upper bound of such an encoder-decoder network trained on a $12GB$ GPU. To increase the output resolution of our proposed network, we keep the basic network architecture used in the depth estimation stage as the core processing component of our color restoration net, as depicted in~\figref{fig:color-module}. Then we wrap the core component with an extra downsampling and upsampling stage. The input image is downsampled using an averaging pooling layer to a resolution of $128\times 128$ and passed through the core process component. At the end of the core component, the output is then upsampled to $512\times 512$ using a deconvolution layer initialized by a bilinear interpolation filter. Two skipping layers are concatenated to preserve high resolution features.
In this way, the main intermediate computation is still done in relatively low resolution. 
We were able to use a batch size of $15$ to train the network on a $12GB$ GPU with this resolution. For both the depth estimation and color correction networks, a Euclidean loss function is used. The pixel values in the images are normalized between $0$ to $1$.

\section{Experimental Setup}\label{s:exp}

We evaluate our proposed method using datasets gathered in both a controlled pure water test tank and from real scientific surveys in the field. As input in-air RGB-D for all experiments, we compile four indoor Kinect datasets (B3DO~\cite{B3DO}, UW RGB-D Object~\cite{UW-dataset}, NYU Depth~\cite{NYU-dataset} and Microsoft 7-scenes~\cite{microsoft-dataset}) for a total of $15000$ RGB-D images.

\subsection{Artificial Testbed}

\begin{figure}
\centering
\subfloat[Rock platform]{
\includegraphics[width=0.23\columnwidth,height=0.27\columnwidth]{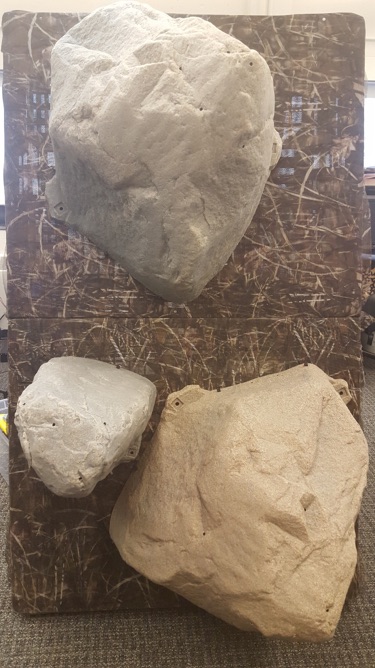}}\hspace{4pt}
\subfloat[Color board]{
\includegraphics[width=0.2\columnwidth,height=0.27\columnwidth]{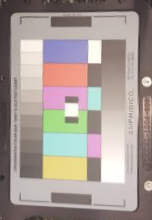}}\hspace{4pt}
\subfloat[MHL test tank]{
\includegraphics[width=0.34\columnwidth,height=0.27\columnwidth]{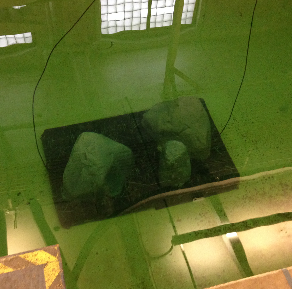}}

\caption{(a) An artificial rock platform and (b) a diving color board are used to provide ground truth for controlled imaging tests.(c) The rock platform is submerged in a pure water test tank for gathering the MHL dataset. \label{fig:rocks-inair}}
\end{figure}

The first survey is done using a 4 ft $\times$ 7 ft man-made rock platform submerged in a pure water test tank at University of Michigan's Marine Hydrodynamics Laboratory (MHL). A color board is attached to the platform for reference (Fig.~\ref{fig:rocks-inair}). 
A total of over 7000 underwater images are compiled from this survey.

 \subsection{Field Tests}

One field dataset was collected in Port Royal, Jamaica, at the site of a submerged city containing both natural and man-made structure. These images were collected with a hand-held diver rig. For our experiments, we compile a dataset consisting of $6500$ images from a single dive. The maximum depth from the seafloor is approximately $1.5m$. 
Another field dataset was collected at a coral reef system near Lizard Island, Australia~\cite{lizard}. The data was gathered with the same diver rig and we assumed a maximum depth of $2.0m$ from the seafloor. We compile a total number of 6083 images from the multi-dive survey within a local area.
 \subsection{Network Training}
For each dataset, we train the WaterGAN network to model a realistic representation of raw underwater images from a specific survey site. The real samples are input to WaterGAN's discriminator network during training, with an equal number of in-air RGB-D pairings input to the generator network. We train WaterGAN on a Titan X (Pascal) with a batch size of 64 images and a learning rate of 0.0002. Through experiments, we found 10 epochs to be sufficient to render realistic images for input to the color correction network for the Port Royal and Lizard Island datasets. We trained for 25 epochs for the MHL dataset. Once a model is trained, we can generate an arbitrary amount of synthetic data. For our experiments, we generate a total of $15000$ rendered underwater images for each model (MHL, Port Royal, and Lizard Island), which corresponds to the total size of our compiled RGB-D dataset.  

Next, we train our proposed color correction network with our generated images and corresponding in-air RGB-D images. We split this set into a training set with $12000$ images and a validation set with $3000$ images.
We train the networks from scratch for both the depth estimation network and image restoration network on a Titan X (Pascal) GPU. For the depth estimation network, we train for $20$ epochs with a batch size of $50$, a base learning rate of $1e^{-6}$, and a momentum of $0.9$.  For the color correction network, we conduct a two-level training strategy. For the first level, the core component is trained with an input resolution of $128\times128$, a batch size of $20$, and a base learning rate of $1e^{-6}$ for 20 epochs. Then we train the whole network at a full resolution of $512\times 512$, with the parameters in core components initialized from the first training step. We train the full resolution model for $10$ epochs with a batch size of $15$ and a base learning rate of $1e^{-7}$.
Results are discussed in \secref{s:res} for all three datasets.

\section{Results and Discussion}\label{s:res}

To evaluate the image restoration performance in real underwater data, we present both qualitative and quantitative analysis for each dataset. We compare our proposed method to image processing approaches that are not range-dependent, including histogram equalization and normalization with the gray world assumption. We also compare our results to a range-dependent approach based on a physical model, the modified Jaffe-McGlamery model (Eqn. 3) with ideal attenuation coefficients~\cite{jordtthesis}. Lastly, we compare our proposed method to Shin et al.'s deep learning approach~\cite{yshin-2016-oceans}, which implicitly models range-dependent information in estimating a transmission map.

\begin{figure*}
\centering
      \captionsetup[subfigure]{justification=centering}
    \subfloat[Raw underwater image][Raw underwater\\image]{
    \includegraphics[width=0.33\columnwidth,height=2.2\columnwidth]{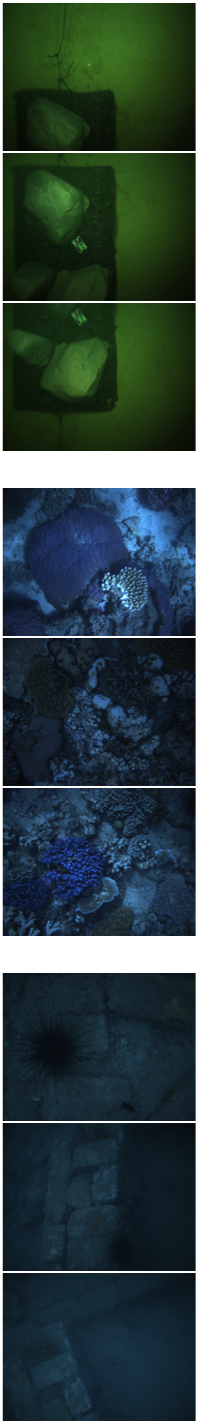}}
    \subfloat[Histogram equalization][Histogram\\equalization]{
    \includegraphics[width=0.33\columnwidth,height=2.2\columnwidth]{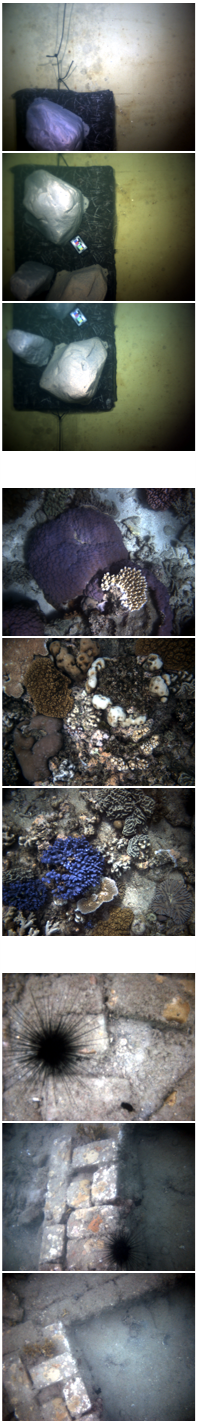}}
    \subfloat[Gray world]{
    \includegraphics[width=0.33\columnwidth,height=2.2\columnwidth]{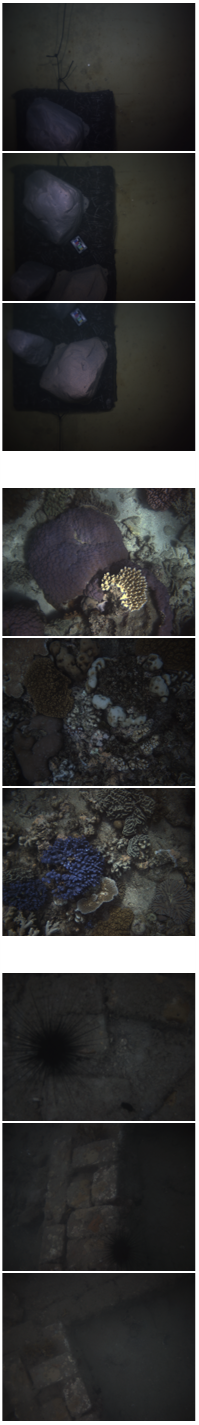}}
    \subfloat[Modified Jaffe-McGlamery][Modified\\Jaffe-McGlamery]{
    \includegraphics[width=0.33\columnwidth,height=2.2\columnwidth]{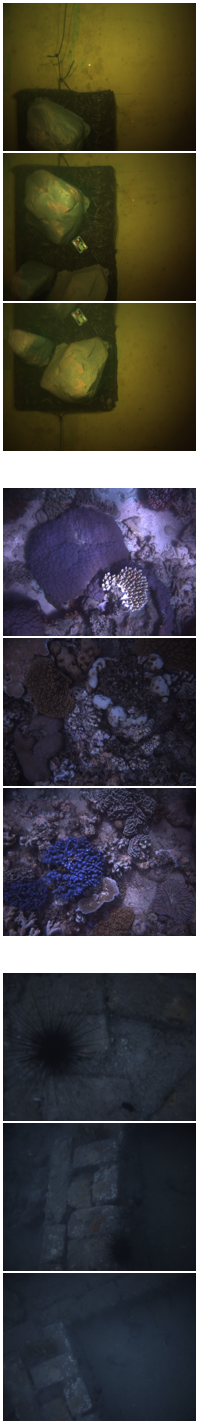}}
    \subfloat[Shin et al.\cite{yshin-2016-oceans}]{
    \includegraphics[width=0.33\columnwidth,height=2.2\columnwidth]{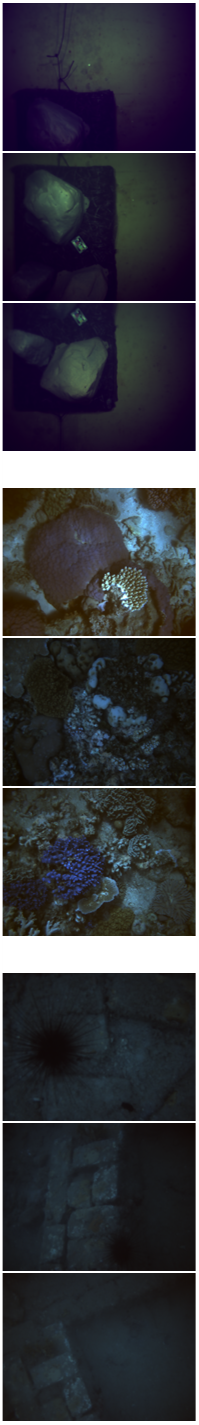}}
    \subfloat[Our Method]{
    \includegraphics[width=0.33\columnwidth,height=2.2\columnwidth]{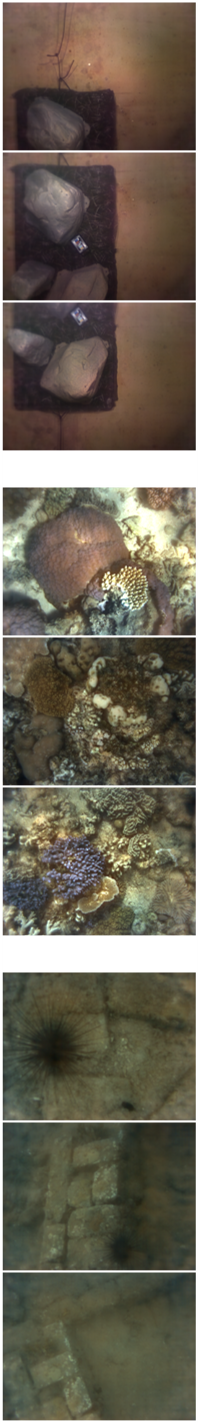}}

\caption{Results showing color correction on the MHL, Lizard Island, and Port Royal datasets (from top to bottom). Each column shows (a) raw underwater images, and corrected images using (b) histogram equalization, (c) normalization with the gray world assumption, (d) a modified Jaffe-McGlamery model (Eqn. 3) with ideal attenuation coefficients, (e) Shin et al.'s deep learning approach, and (f) our proposed method.\label{fig:grid-result}
}

\end{figure*}

Qualitative results are given in Figure~\ref{fig:grid-result}. Histogram equalization looks visually appealing, but it has no knowledge of range-dependent effects so corrected color of the same object viewed from different viewpoints appears with different colors. Our proposed method shows more consistent color across varying views, with reduced effects of vignetting and attenuation compared to the other methods. We demonstrate these findings across the full datasets in our following quantitative evaluation.


We present two quantitative metrics for evaluating the performance of our color correction: color accuracy and color consistency. For accuracy, we refer to the color board attached to the submerged rock platform in the MHL dataset. Table~\ref{t:accuracy} shows the Euclidean distance of intensity-normalized color in RGB-space for each color patch on the color board compared to an image of the color board in air. These results show that our method has the lowest error for blue, red, and magenta. Histogram equalization has the lowest error for cyan, yellow and green recovery, but our method still outperforms the remaining methods for cyan and yellow.

\vspace{-1mm}
\begin{table}[h]
\caption{Color correction accuracy based on Euclidean distance of intensity-normalized color in RGB-space for each method compared to the ground truth in-air color board.\vspace{-2mm}}
\label{t:accuracy}
\begin{center}
\begin{tabular}{|c|c|c|c|c|c|c|}
\hline
 & \textbf{Raw} & \textbf{\shortstack{Hist.\\Eq.}} & \textbf{\shortstack{Gray\\World}} & \textbf{\shortstack{Mod.\\J-M}} & \textbf{\shortstack{Shin\cite{yshin-2016-oceans}}} & \textbf{\shortstack{Prop.\\Meth.}}\\ \hline
 Blue & 0.3349 &	0.2247 &	0.2678	& 0.2748 &	0.1933 &	\textbf{0.1431}\\ \hline
 Red & 0.2812	& 0.0695 &	0.1657 &	0.2249	& 0.1946	& \textbf{0.0484}\\ \hline
 Mag. & 0.3475	& 0.1140 &	0.2020 &	0.298 &	0.1579 &	\textbf{0.0580}\\ \hline
 Green & 0.3332	& \textbf{0.1158}	& 0.1836	& 0.2209 &	0.2013	& 0.2132\\ \hline
 Cyan & 0.3808	& \textbf{0.0096}	& 0.1488	& 0.3340	& 0.2216 &	0.0743\\ \hline
 Yellow & 0.3599 &	\textbf{0.0431}	& 0.1102	& 0.2265 &	0.2323	& 0.1033\\ \hline 

\end{tabular}
\end{center}
\end{table}
\vspace{-4mm}

To analyze color consistency quantitatively, we compute the variance of intensity-normalized pixel color for each scene point that is viewed across multiple images. Table~\ref{t:consistency} shows the mean variance of these points. Our proposed method shows the lowest variance across each color channel. This consistency can also be seen qualitatively in Fig.~\ref{fig:grid-result}.

\vspace{-1mm}
\begin{table}[h]
\caption{Variance of intensity-normalized color of single scene points imaged from different viewpoints.\vspace{-2mm}}
\label{t:consistency}
\begin{center}
\begin{tabular}{|c|c|c|c|c|c|c|}
\hline
 & \textbf{Raw} & \textbf{\shortstack{Hist.\\Eq.}} & \textbf{\shortstack{Gray\\World}} & \textbf{\shortstack{Mod.\\J-M}} & \textbf{\shortstack{Shin\cite{yshin-2016-oceans}}} & \textbf{\shortstack{Prop.\\Meth.}}\\ \hline
 Red & 0.0073 &	0.0029	& 0.0039	& 0.0014	& 0.0019	& \textbf{0.0005}\\ \hline
 Green & 0.0011	& 0.0021 &	0.0053	& 0.0019	& 0.0170	& \textbf{0.0007}\\ \hline
 Blue & 0.0093	& 0.0051	& 0.0042	& 0.0027 &	0.0038	& \textbf{0.0006}\\ \hline

\end{tabular}
\end{center}
\end{table}
\vspace{-4mm}

We also validate the trained network on the testing set of synthetic data and the validation results are given in Table~\ref{t:validation}. We use RMSE as the error metric for both color and depth. These results show that the trained network is able to invert the model encoded by the generator.


\vspace{-1mm}
\begin{table}[htbp]
\centering
\caption{Validation error in pixel value is given in RMSE in RGB-space. Validation error in depth is given in RMSE (m).\vspace{-2mm}}
\label{t:validation}
\begin{tabular}{|c|c|c|c|c|}
\hline
     \textbf{Dataset} & \textbf{Red} & \textbf{Green} & \textbf{Blue}  &\shortstack{\textbf{Depth}\\\textbf{RMSE}}\\
 \hline
     Synth. MHL & 0.052 & 0.033& 0.055 &0.127\\
 \hline
     Synth. Port Royal & 0.060 & 0.041& 0.031 &0.122\\
\hline
   Synth. Lizard& 0.068 & 0.045& 0.035 &0.103\\
   \hline 
\end{tabular}
\end{table}
\vspace{-4mm}

In terms of the computational efficiency, the forward propagation for depth estimation takes $0.007s$ on average and the color correction module takes $0.06s$ on average, which is efficient for real-time applications.

It is important to note that our depth estimation network recovers accurate relative depth, not necessarily absolute depth. This is due to the scale ambiguity inherent to the monocular depth estimation problem. 
To evaluate the depth estimation in real underwater images, we compare our estimated depth with depth reconstructed from stereo images available for the MHL dataset in a normalized manner, ignoring the pixels where no depth is recovered from stereo reconstruction due to lack of overlap or feature sparsity. The RMSE of normalized estimated depth and the normalized stereo reconstructed depth is $0.11m$.

\begin{figure}
\centering
 \subfloat[Raw image patch]{
    \includegraphics[width=0.3\columnwidth]{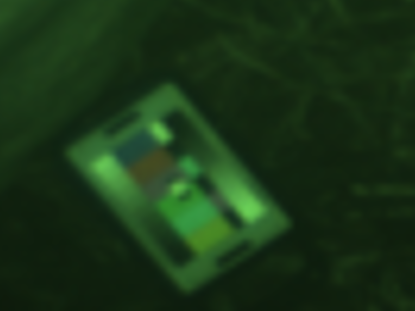}}
    \subfloat[Restored image without skipping layers]{
    \includegraphics[width=0.3\columnwidth]{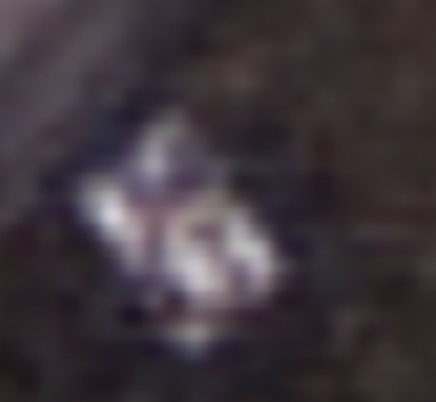}}
    \subfloat[Proposed output]{
    \includegraphics[width=0.3\columnwidth]{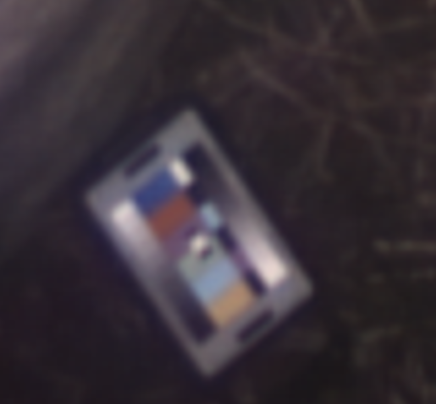}}
\caption{Zoomed-in comparison of color correction results of an image with and without skipping layers.\label{fig:detail}}
\end{figure}

To evaluate the improvement in image quality due to skipping layers in the color correction network, we train the network at the same resolution with and without skipping layers. 
For the first pass of core component training, the network without skipping layers takes around $30$ epochs to reach a stable loss, while the proposed network with skipping layers takes around $15$ epochs. The same trend holds for full model training, taking $10$ and $5$ epochs, respectively. Figure~\ref{fig:detail} shows a comparison of image patches recovered from both versions of the network. This demonstrates that using skipping layers helps to preserve high frequency information from the input image.

One limitation of our model is in the parameterization of the vignetting model, which assumes a centered vignetting pattern. This is not a valid assumption for the MHL dataset, so our restored images still show some vignetting though it is partially corrected. These results could be improved by adding a parameter that adjusts the center position of the vignetting pattern over the image. This demonstrates a limitation of augmented generators, more generally. Since they are limited by the choice of augmentation functions, augmented generators may not fully capture all aspects of a complex nonlinear model~\cite{RenderGAN}. We introduce a convolutional layer into our augmented generator that is meant to capture scattering, but we would like to experiment with adding additional layers to this stage for capturing more complex effects, such as lighting patterns from sunlight in shallow water surveys. To further increase the network robustness and enable the generalization to more application scenarios, we would also like to train our network across more datasets covering a larger variety of environmental conditions including differing illumination and turbidity. 

Source code, sample datasets, and pretrained models are available at \url{https://github.com/kskin/WaterGAN}.

\section{Conclusions}\label{s:con}
This paper proposed WaterGAN, a generative network for modeling underwater images from RGB-D in air. We showed a novel generator network structure that incorporates the process of underwater image formation to generate high resolution output images. We then adapted a dense pixel-wise model learning pipeline for the task of color correction of monocular underwater images trained on RGB-D pairs and corresponding generated images. We evaluated our method on both controlled and field data to show qualitatively and quantitatively that our output is accurate and consistent across varying viewpoints. 
 There are several promising directions for future work to extend this network. Here we train WaterGAN and the color correction network separately to simplify initial development of our methods. Combining these networks into a single network to allow joint training would be a more elegant approach. Additionally, this would allow the output of the color correction network to directly influence the WaterGAN network, perhaps enabling development of a more descriptive loss function for the specific application of image restoration. 



\section*{ACKNOWLEDGMENTS}

The authors would like to thank the Australian Centre for Field Robotics for providing the Lizard Island dataset, the Marine Hydrodynamics Laboratory at the University of Michigan for providing access to testing facilities, and Y.S. Shin for sharing source code. This work was supported in part by the National Science Foundation under Award Number: 1452793, the Office of Naval Research under award N00014-16-1-2102, and by the National Oceanic and Atmospheric Administration under award NA14OAR0110265.

\renewcommand{\bibfont}{\normalfont\small}
\printbibliography

\onecolumn

\end{document}